\documentclass{article}

\usepackage{microtype}
\usepackage{graphicx}
\usepackage{subfigure}
\usepackage{booktabs} 

\usepackage{hyperref}

\usepackage[accepted]{icml2024}

\usepackage{amsmath}
\usepackage{amssymb}
\usepackage{mathtools}
\usepackage{amsthm}

\usepackage{booktabs} 
\usepackage{array}    
\usepackage{siunitx}  
\usepackage{subcaption}
\usepackage{multirow}
\usepackage{color, colortbl}
\definecolor{LightCyan}{rgb}{0.94,1,1}
\definecolor{LightRed}{rgb}{1,0.94,0.94}
\definecolor{Gray}{rgb}{0.8,0.8,0.8}
\definecolor{LightGray}{rgb}{0.92,0.92,0.92}
\definecolor{LLightGray}{rgb}{0.97,0.97,0.97}
\definecolor{mygreen}{rgb}{0.4,0.7,0.306}
\definecolor{myblue}{rgb}{0.294,0.447,0.796}
\usepackage{hhline}
\usepackage{nicematrix}

\usepackage[capitalize,noabbrev]{cleveref}

\theoremstyle{plain}

\theoremstyle{definition}

\theoremstyle{remark}

\usepackage[textsize=tiny]{todonotes}

\usepackage{algorithmic}
\usepackage{algorithm}

\usepackage{comment}

\begin{document}

\twocolumn[
\icmltitle{Discriminative Adversarial Unlearning}




\begin{icmlauthorlist}
\icmlauthor{Rohan Sharma}{yyy}
\icmlauthor{Shijie Zhou}{yyy}
\icmlauthor{Kaiyi Ji}{yyy}
\icmlauthor{Changyou Chen}{yyy}
\end{icmlauthorlist}

\icmlaffiliation{yyy}{Department of CSE, University at Buffalo}

\icmlcorrespondingauthor{Rohan Sharma}{rohanjag@buffalo.edu}

\icmlkeywords{Machine Learning, ICML}

\vskip 0.3in
]



\printAffiliationsAndNotice{}  





\begin{abstract}
We introduce a novel machine unlearning framework founded upon the established principles of the min-max optimization paradigm. We capitalize on the capabilities of strong Membership Inference Attacks (MIA) to facilitate the unlearning of specific samples from a trained model. We consider the scenario of two networks, the attacker $\mathbf{A}$ and the trained defender $\mathbf{D}$ pitted against each other in an adversarial objective, wherein the attacker aims at teasing out the information of the data to be unlearned in order to infer membership, and the defender unlearns to defend the network against the attack, whilst preserving its general performance. The algorithm can be trained end-to-end using backpropagation, following the well known iterative min-max approach in updating the attacker and the defender. We additionally incorporate a self-supervised objective effectively addressing the feature space discrepancies between the forget set and the validation set, enhancing unlearning performance. Our proposed algorithm closely approximates the ideal benchmark of retraining from scratch for both random sample forgetting and class-wise forgetting schemes on standard machine-unlearning datasets. Specifically, on the class unlearning scheme, the method demonstrates near-optimal performance and comprehensively overcomes known methods over the random sample forgetting scheme across all metrics and multiple network pruning strategies. 
Code is available at: \href{https://github.com/rohan1561/Gunlearn}{Link}.
\end{abstract}

\section{Introduction}
With over \$4 billion paid in settlements over privacy concerns by Big Tech firms since the enforcement of the General Data Protection Regulation \cite{EuropeanParliament2016a} in 2018 \footnote{Source: \href{https://dataprivacymanager.net/5-biggest-gdpr-fines-so-far-2020/}{Data Privacy Manager}}, and more than 70\% of citizens from nine major countries expressing support for increased governmental intervention in Big Tech\footnote{Source: \href{https://www.amnesty.org/en/latest/press-release/2019/12/big-tech-privacy-poll-shows-people-worried/}{Amnesty International}}, significant concerns arise regarding individual privacy and the potential erosion of trust between human users and AI systems. Owing to their widespread incorporation, digital systems now function as vast repositories of personal data, creating a comprehensive digital footprint that reflects a broad spectrum of online behaviors, interactions, and communication patterns \cite{nguyen2019debunking}. This footprint is evident across various forms of user-generated content, including product reviews, blog posts, social media activities, and contributions to collaborative platforms such as Wikipedia \cite{nguyen2021judo, carpentier2021poster}. The application of artificial intelligence (AI) in areas such as targeted advertising and personalized product placement, along with the emergence of potentially harmful technologies like deepfakes, further underscores a critical concern regarding the lack of transparency and control over personal data, highlighting the need for protective measures in the management and utilization of personal information \cite{yevseiev2021modeling, al2023towards}. The recent legal developments echo these concerns with a notable development being, "the right to be forgotten," \cite{dang2021right, tankard2016gdpr} which requires the deletion of personal data from both digital databases and machine learning models at an individual's request. However, retraining AI models to accommodate each deletion request is impractical, given the potential cost of millions of dollars, coupled with substantial environmental impact.

In response to these challenges, there has been a notable surge in the development of algorithmic unlearning strategies \cite{wang2023kga, zhang2023machine, Hoang_2024_WACV, chen2023boundary, lin2023erm, liu2023muter, che2023fast}. These strategies aim to modify existing models, allowing them to selectively 'unlearn' specific samples from their training data. The approaches offer viable alternatives to complete retraining from scratch, providing cost-effective solutions that enhance data security and privacy while preserving the efficacy of the models. Despite their significant effectiveness, a majority of these measures \cite{liu2023muter, mehta2022deep, gandikota2023erasing, heng2023continual, heng2023selective} primarily address the problem in the parametric space. They introduce modifications within the pre-training objectives \cite{lin2023erm} or the pre-processing phase \cite{liu2017identification, garcia2020sensitive} of trained models, often entailing performance-cost trade-offs due to the evaluation of hessian approximations or proposals for complete overhauls of training procedures. 

Membership Inference Attacks (MIA), on the other hand as a field of study, has concurrently bloomed \cite{7958568, salem2018mlleaks, nasr2018machine, 255348, song2021systematic, yuan2022membership} exposing significant privacy related threats to trained models. For a trained model, MIAs are tasked with identification of inference of patterns in the model's behaviour due to samples arising from its training set and other sets. Using these patterns, a successful attacker may avail themselves with personally identifying information \cite{veale2018algorithms, fredrikson2015model, tramer2016stealing, ganju2018property} posing significant threats to the use of AI towards general welfare. 

We posit that the objectives of the MIA and machine unlearning may be formulated as adversarially inclined and one may harness the recent developments in the fields in order to orchestrate a min-max framework and successfully conduct machine unlearning in pretrained models, preserving our trust in machine learning as a service (MLaaS). To our knowledge, our contribution is the first to formulate the problem of unlearning as an adversarial task wherein two networks, the attacker and the defender engage in a game against each other, which consequently leads towards efficient and effective unlearning measures learned in the defender, whilst notable retention of it's original capacity for classification. We summarize our contribution in this work as follows.
\begin{itemize}
    \item We introduce a pioneering framework for machine unlearning, employing a min-max optimization procedure that engages neural networks in an end-to-end trainable game through backpropagation. Importantly, this framework eliminates the necessity for unrolling or approximation in the optimization procedures of the networks, and avoids the need for explicit gradient manipulation.
    \item We innovate by repurposing the Barlow Twins objective \cite{pmlr-v139-zbontar21a} for unlearning, yielding a substantial performance boost. This repurposing opens avenues for applying self-supervised learning research to provide practical solutions in machine unlearning.
    \item Through extensive experimentation, we showcase the effectiveness of our framework, achieving near-optimal performance and setting new benchmarks for machine unlearning on the standard machine-unlearning Cifar-10 and Cifar-100 datasets. Our framework excels under both random and class-wise forgetting scenarios.
\end{itemize}
\section{Related Work}
\subsection{Machine Unlearning}
Machine unlearning endeavors to mitigate the influence of specific data points or classes from a trained ML model, primarily addressing privacy concerns in accordance with regulations \cite{rosen2011right, hoofnagle2019european}. Recent developments in the field has led to a taxonomy for methods broadly classifying the objectives as follows.
\paragraph{Exact unlearning} Retraining the model from the scratch on the data filtering out forgetting dataset is the most straightforward and effective manner to update ML models for unlearning, ensuring complete deletion of undesirable samples or aspects from a pre-trained model. \citet{7163042} explore this mode of unlearning over Naive Bayes classifiers whereas \citet{NEURIPS2019_cb79f8fa} implement deletion algorithms for $k\text{-}means$ clustering, rendering them unscalable to deep neural networks. Additionally, recent efforts \cite{bourtoule2021machine} also encourage training on shards of data, subsequently incorporating aggregation, effectively alleviating the expensive nature of training a large model from scratch. However, the measures signify impracticality in the face of large queries for deletion, requiring inordinate amounts of compute especially for models wherein the training cost estimates exceed millions of dollars.

\begin{figure*}[ht]
\vskip 0.2in
\begin{center}
\centerline{\includegraphics[width=1.25\columnwidth]{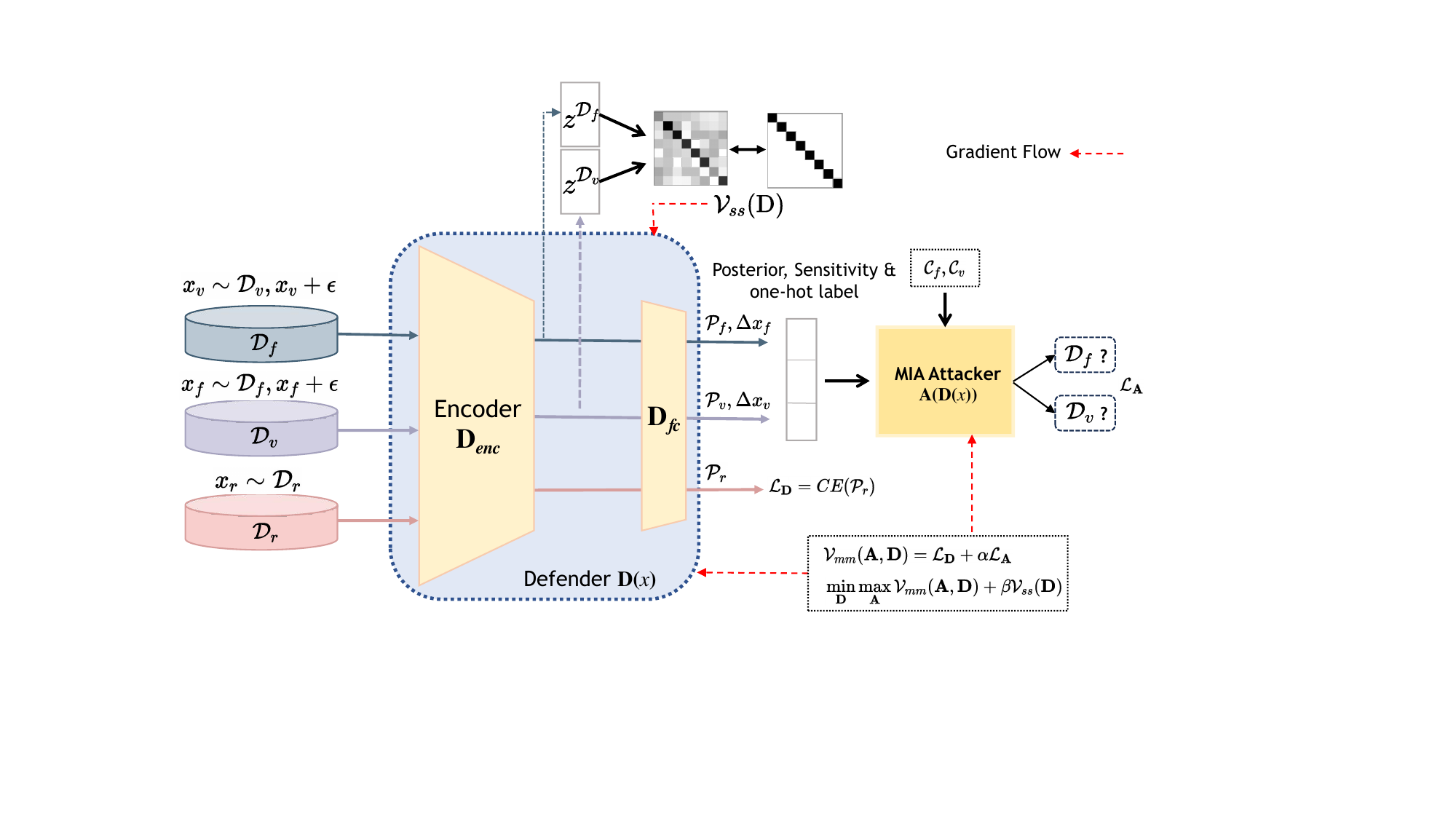}}
\caption{\textbf{Overview of the proposed framework.} The figure depicts the interplay between the attacker and defender networks. The defender provides output and sensitivity information to the attacker which in turn provides feedback to the defender for unlearning. The objective is supplemented with a feature space self supervised regularization between the forget and validation sets.}
\label{fig:model}
\end{center}
\vskip -0.2in
\end{figure*}

\paragraph{Approximate unlearning} Acknowledgement of the said infeasibility of training large models from scratch has therefore led to an upsurge in palliative measures to selectively eradicate information from trained models. Model finetuning as an approach \cite{warnecke2021machine, golatkar2020eternal} aims to finetune the model on the retain set of data to induce catastrophic forgetting within trained models. \citet{graves2021amnesiac, thudi2022unrolling} on the other hand propose retraining the model on corrupted labels for sensitive data providing for a relatively cost-effective measure to induce unlearning. Hessian matrix approximation largely characterizes some of the works \cite{mehta2022deep}. Use of fisher information matrices \cite{becker2022evaluating, liu2023unlearning} allows for a direct albeit expensive estimation of the expectation of the hessian due to $\log p(y|x, \theta)$, allowing for strategies pertinent to noising/masking if the said estimation for the data to be forgotten. Fisher forgetting \cite{becker2022evaluating, liu2023unlearning} broadly aims at a similar estimation of the hessians using FIMs in order to conduct informed corruption of samples in the image space. Model decisions on the other hand may lend to some avenues for manipulation \cite{chen2023boundary, wang2023kga} in order to induce forgetting using statistical distance based metrics such as $KL$ divergence. Unlearning measures have also experienced a surge in interest within the generative models space and in Large Language Models (LLMs). In the generative model domain \cite{heng2023continual, heng2023selective, gandikota2023erasing}, efforts are directed towards inhibiting diffusion or Generative Adversarial Network (GAN)-based models from generating questionable content. Model editing methods \cite{DBLP:journals/corr/abs-2104-08696, DBLP:journals/corr/abs-2012-14913, NEURIPS2022_6f1d43d5} aim to edit factual information encoded within pre-trained LLMs. However, these approaches predominantly focus on selective influential aspects of the trained model in isolation, overlooking the potential for incorporating separate networks to expedite the unlearning process in the model under consideration. Furthermore, in order to incorporate expensive measures for estimation of second order information, significant approximations are formulated which determine the efficiency and performances of the methods.

\subsection{Membership Inference Attacks (MIA)}
The purpose of a MIA is to uncover from a trained model (target), information revealing whether a sample was involved in the training process of the target model. Assuming black box conditions, the attack model trains itself to make binary predictions using the outputs of the target, inferring belonging or lack thereof of samples in the training set of target. This formulation was first proposed by \citet{7958568}, wherein they create multiple shadow models to replicate the behavior of a target model. The shadow models are then utilized to generate data for training a binary classifier to determine membership. Building upon this concept, \citet{salem2018mlleaks} enhanced the efficacy of the attack by employing a single shadow model. Refinements to the method were achieved through the integration of supplementary discriminative information \cite{nasr2018machine}, which included considerations of labels. Additional contributions \cite{255348, yeom2018privacy, song2019privacy, song2021systematic} introduced evaluative metrics where the membership status of a record is directly ascertained by comparing predefined thresholds based on metrics such as prediction confidences, entropy, or modified entropy of the record. Song and Mittal demonstrated that, by establishing a class-dependent threshold, the metric-based classifier could achieve comparable or even superior inference performance compared to the neural network-based classifier \cite{song2021systematic}. In this work we adopt the recent work of \citet{yuan2022membership} towards proposing an effective mechanism for MIA on pruned models. Briefly, the work proposes a transformer based architecture that incorporates multiple sources of information about the target model's inferential patterns in order to effectively conduct MIA against the model.

Leveraging the advancements in these fields we propose a union of the two, formulating an adversarial game between the defender network, tasked with orchestrating the unlearning process and the attacker network leveraging MIA to infer the membership status of the data to be unlearned. Therefore, under the context of this adversarial framework, the successful expunging of a data point's influence on the model is indicated when the attacker is unable to distinguish whether the data point originated from the training set or is a representative instance of unseen data.

\section{The Proposed Method}
Let $\mathcal{D}$ represent the set of training samples utilized for training a network with parameters $\theta_0$. The sets $\mathcal{D}_v$ and $\mathcal{D}_t$ denote the validation and test sets, respectively, employed for evaluating the model, comprising samples that were not part of the training process. $\mathcal{D}_f \subseteq \mathcal{D}$ is defined as the subset of samples targeted for removal from the model, resulting in updated parameters $\theta_u$. The set $\mathcal{D}_r$ constitutes the retained samples, defined as $\mathcal{D}_r = \mathcal{D} \setminus \mathcal{D}_f$. Two primary scenarios of forgetting are considered: random forgetting, where the model is required to forget samples randomly selected from $\mathcal{D}$, and class-wise forgetting, where samples specific to a certain class are targeted for removal. The benchmark for evaluating an unlearning algorithm is the model trained solely on the retained set. Specific to our method, we introduce in our framework, an attacker model $\mathbf{A}$ trained to infer the membership of a given sample.

From the standpoint of $\mathbf{A}$, in order for a trained defender $\mathbf{D}$ to forget its training due to $\mathcal{D}_f$, the attacker must be unable to distinguish between its predictions for $\mathcal{D}_f$ and $\mathcal{D}_v$. This motivates the algorithm of unlearning wherein $\mathbf{A}$ and $\mathbf{D}$ are trained adversarially with their objectives pit against each other, essentially approaching an optimal convergence, wherein the performance of the gold standard method of training from scratch on $\mathcal{D}_r$ is recovered by the defender and the attacker predictions equal $0.5$ for all samples arising from $\mathcal{D}_f$ and $\mathcal{D}_v$. In what follows, we delve into the specifics of the attacker and defender architectures and objectives and describe our formulation of the problem that allows us to leverage a these networks in the min-max framework, in order to enable unlearning. Following this, we elucidate the integration of the self-supervised objective, enhancing the method's efficacy and contributing to the overall objective that defines the algorithm.

\subsection{The Attacker} \label{sec:attacker}

Consider a target defender model $\mathbf{D}$, which is trained on  the training set $\mathcal{D}$. The objective of the attacker model $\mathbf{A}$, is to deduce a binary prediction indicating the presence or absence of a specific sample in the training set. Typically, the scenario is characterized as a black-box setting, where the attacker's access is constrained to the outputs of the defender model. The attacker model is trained by utilizing the defender's inferences on samples originating from both the set of forgotten samples $\mathcal{D}_f$ and the validation set $\mathcal{D}_v$, with the distinction that the former was used in the training process of the defender, while the latter was not. As we shall see later, it is strategically advantageous to employ a potent attacker that assimilates substantial information to achieve accurate inferences regarding membership. Accordingly, we consider the recent formulation proposed by \citet{yuan2022membership}. Briefly, this attacker is designed to process three crucial pieces of information for each sample:

\begin{itemize}
    \item The output predictions tensor generated by the defender for the given sample.
    \item The sensitivity of the defender to noisy inputs, evaluated as $\Delta(x) = \frac{1}{n} \sum_{i=1}^{n} \left\| \mathbf{D}(x) - \mathbf{D}(x + \mathbf{\epsilon}_i) \right\|$ where $\epsilon \sim \mathcal{N}(0, 1)$. We evaluate the sensitivity tensor upon averaging over $n=10$ noisy samples.
    \item The one-hot label information $y$ of the sample, indicating the class information.
\end{itemize}
The attacker then formulates the informed inputs upon concatenating these tensors $\mathbf{D}(x_f)\prime = [\mathbf{D}(x_f), \Delta(x_f), y_f ]$, $\mathbf{D}(x_v)\prime = [\mathbf{D}(x_v), \Delta(x_v), y_v]$, and further encodes using a fully connected (FC) layer. The resulting representation is then propagated through a transformer architecture comprising self-attention layers eventually allowing for a binary prediction score. Through this process, the attacker aims to tease out the influence of the forget set $\mathcal{D}_f$ within the defender model and accordingly maximizes the log-likelihood given by $\mathcal{L}_{\mathbf{A}} = \mathbb{E}_{x_f\sim\mathcal{D}_f, x_v\sim\mathcal{D}_v}[\log(\mathbf{A}(\mathbf{D}(x_f)\prime)+\log(1-\mathbf{A}(\mathbf{D}(x_v)\prime)]$.  

\subsection{The Defender}\label{sec:defender}
A network trained on the dataset $\mathcal{D}$ is now tasked with retaining its performance on $\mathcal{D}_v$ and $\mathcal{D}_t$ whilst forgetting its training on $\mathcal{D}_f$. The former may be accomplished using the supervised classification objective wherein the network is provided with labeled samples originating from the $\mathcal{D}_r$. This is typically implemented as the cross entropy loss $\mathcal{L}_{\mathbf{D}} = \mathbb{E}_{x_r\sim\mathcal{D}_r}[-\sum_{i=1}^{K} y_{ri} \cdot \log(\mathbf{D}(x_r)_i)]$ for a $K$ class classification problem, which $\mathbf{D}$ aims to minimize. For the latter however, the defender may leverage the capacity of the attacker network to distinguish between its outputs from $\mathcal{D}_f$ and $\mathcal{D}_v$. Accordingly, in order to effectively forget, the defender network also minimizes the objective of the attacker, therefore defending against an active MIA conducted by the attacker.

\subsection{Adversarial Unlearning}
The attacker and defender networks are now entangled in adversarial objectives, wherein the defender network not only seeks to defend against the attacker, but also aims to preserve its performance on the validation set $\mathcal{D}_v$ and the test set $\mathcal{D}_t$ while the attacker is tasked with conducting MIA against the defender. We may therefore formalize the interaction between the attacker and the defender as a min-max game, defined as follows:
\begin{align}
\min_\mathbf{D}\max_\mathbf{A}\mathcal{V}_{mm}(\mathbf{A}, &\mathbf{D}) \triangleq \mathcal{L}_{\mathbf{D}} + \alpha \mathcal{L}_{\mathbf{A}} \nonumber \\ 
&= \mathbb{E}_{x_r\sim\mathcal{D}_r}[-\sum_{i=1}^{K} y_{ri} \cdot \log(\mathbf{D}(x_r)_i)] \nonumber \\
&+ \alpha[\mathbb{E}_{x_f\sim\mathcal{D}_f, x_v\sim\mathcal{D}_v}[\log(\mathbf{A}(\mathbf{D}(x_f)\prime)\nonumber\\
&+ \log(1-\mathbf{A}(\mathbf{D}(x_v)\prime))]],
\end{align}
where $\alpha$ modulates the influence of the attacker's objective on the defender network which also optimizes for a $K$ class classification problem. To facilitate the training of the networks with the specified objective, we employ the iterative updating scheme common to the min-max optimization frameworks for deep network learning \cite{goodfellow2020generative, arjovsky2017wasserstein, gulrajani2017improved}, for both the defender and the attacker. The trained defender is provided with samples from $\mathcal{D}_f$ and $\mathcal{D}_v$ on which it conducts inference. The attacker subsequently uses these inferences to generate its inputs, as outlined in Section \ref{sec:attacker}. During this process, the attacker acquires knowledge of the inference patterns exhibited by the defender and endeavors to distinguish between the defender's outputs for samples from $\mathcal{D}_f$ and $\mathcal{D}_v$, by maximizing a binary classification objective. This discrimination task is relatively easier for the attacker in the early stages of the algorithm. As the training progresses, the defender becomes aware of the attacker's motives and consequently adjusts its own learning objectives to defend against the attack, minimizing the attacker's objective, leading to a decline in its performance on the set of retained samples $\mathcal{D}_r$. In response to this degradation, the defender optimizes for learning to classify on $\mathcal{D}_r$ to preserve its general capacity for classification on $\mathcal{D}$. This adaptive learning process reflects the dynamic interplay between the defender and the attacker during training, ultimately influencing the defender's strategies to balance classification performance on the retained set and resistance against the MIA.

One may however consider making practical adjustments to the objective, in consideration of the problem of weaker gradients as identified in the work of 
 Generative Adversatial Nets (GAN) \citet{goodfellow2020generative} and instead maximize $\mathbb{E}_{x_f\sim\mathcal{D}_f, x_v\sim\mathcal{D}_v}[\log(\mathbf{A}(1-\mathbf{D}(x_f)) + \log(\mathbf{A}(\mathbf{D}(x_v)))]$ to empirically adjust for the differential capacities of the attacker and the defender networks to counter each other's efforts. Additionally, one may also incorporate improvements over the GAN algorithm \cite{arjovsky2017wasserstein, gulrajani2017improved, kodali2017convergence} in order to ensure better convergence. Incorporation of objectives pertinent to unlearning such as gradient ascent using a negative of the loss function for $\mathcal{D}_f$ \cite{graves2021amnesiac, thudi2022unrolling}, or fisher forgetting through additive gaussian noise \cite{becker2022evaluating, izzo2021approximate} is also feasible through relatively nominal adjustments to the algorithm.

\subsection{Self-supervised Regularization} \label{sec:ssr}
The min-max objective strives to render the outputs of $\mathbf{D}$ indistinguishable to a proficient attacker within the output space. Nevertheless, the defender network tends to overcompensate by emphasizing exclusively on the final output layer, aiming to enhance its competence against the attacker, and the feature space discrepancies between $\mathcal{D}_f$ and $\mathcal{D}_v$ remain overlooked. In order to address these discrepancies we introduce a self supervised regularization term that enables the defender to inherit the task of forgetting samples within the network's parameters and eases the burden on the final layer of the network. We motivate this objective through the notion that the features generated for the forget set $\mathcal{D}_f$ must on average be similar to those arising from the validation set $\mathcal{D}_v$ and therefore the test set $\mathcal{D}_t$. In our application we consider the feature space redundancy reduction principle utilized by \citet{pmlr-v139-zbontar21a} that aims to enforce invariance between features of samples that in principle must be similar. The defender $\mathbf{D}$ essentially consisting of the encoder and the FC layer $\mathbf{D}(.) = \mathbf{D}_{fc}(\mathbf{D}_{enc}(.))$, upon processing the input samples, outputs mean-centered features $\{z_b: z_b=\mathbf{D}_{enc}(x_b), z_b \in \mathbb{R}^D\}$, where $b$ indexes the batch. Subsequently, for a cross-correlation matrix $\mathcal{C}$ computed between features from $\mathcal{D}_f$ and $\mathcal{D}_v$, defined as, 
\begin{equation*}
    \mathcal{C}_{ij} \triangleq \frac{\sum_b z_{b,i}^{\mathcal{D}_f} z_{b,j}^{\mathcal{D}_v}}{\sqrt{\sum_b (z_{b,i}^{\mathcal{D}_f})^2}\sqrt{\sum_b (z_{b,j}^{\mathcal{D}_t})^2}}
\end{equation*}
    
the self supervised regularization is formulated as follows:
    
\begin{equation} \label{eqn:ssreg}
    \min_D \mathcal{V}_{ss}(\mathbf{D}) \triangleq  \sum_i(1 - \mathcal{C}_{ii})^2 + \lambda \sum_i\sum_{j \neq i}\mathcal{C}_{ij}^2,
\end{equation}
where $\lambda$ was set to $5e\text{-}3$ as recommended in \citet{pmlr-v139-zbontar21a}.
The regularization plays a crucial role in augmenting the overall performance of the algorithm, as elucidated through ablation studies in Appendix Section \ref{sec:ssr_ablations}. However, one may surmise several other variants of the self-supervised objective \cite{DBLP:journals/corr/abs-2105-04906, Caron_2021_ICCV}. Additionally, contrastive objectives \cite{https://doi.org/10.48550/arxiv.2002.05709, anonymous2024auccl, dwibedi2021little} that enable similarity between $\mathcal{D}_f$ and $\mathcal{D}_v$ and distance between $\mathcal{D}_f$ and $\mathcal{D}_r$, may also be incorporated as regularizers. One may also leverage class label information to motivate self-supervised objectives within the classes that the samples belong to. Furthermore, one may incorporate a separate projector network over the defender in order to implement this regularization as is commonly practiced in SSL literature.
The overall objective that we utilize in order to train the framework through gradient descent is formulated as:

\begin{equation}
    \min_\mathbf{D}\max_\mathbf{A}\mathcal{V}(\mathbf{A}, \mathbf{D}) \triangleq \mathcal{V}_{mm}(\mathbf{A}, \mathbf{D}) + \beta \mathcal{V}_{ss}(\mathbf{D}),
\end{equation}
which concludes our method. Here, $\beta$ signifies the strength of regularization which is empirically determined. The complete procedure for this approach is listed Algorithm \ref{alg:main}, which delineates the iterative update schema used to implement the min-max approach, sequentially updating the attacker and the defender networks, leading to the unlearned defender network $\mathbf{D}_{\theta_u}$. An illustration of the procedure is provided under Figure \ref{fig:model}.

\begin{algorithm*} 
\caption{Adversarial Unlearning}
\label{alg:main}
\begin{algorithmic}[1]
\STATE \textbf{Input:} $\mathcal{D}_f$, $\mathcal{D}_v$, $\mathcal{D}_r$, $\mathbf{A}_{\theta_{\mathbf{A}}}$, $\mathbf{D}_{\theta_{\mathbf{D}}}$
\STATE \textbf{Parameters:} Batch Size $B$, Learning Rates $\eta_{\mathbf{A}}$, $\eta_{\mathbf{D}}$, Parameters $\alpha, \beta, \lambda$
\FOR{epoch}
    \STATE Sample $(x_r, y_r)\sim\mathcal{D}_r$
    \STATE Evaluate $\mathbf{D}(x_f)\prime\,\forall x_f\in\mathcal{D}_f$, $\mathbf{D}(x_v)\prime\,\forall x_v\in\mathcal{D}_v$
    \STATE Evaluate $\mathcal{V}_{ss}\,\forall (x_f, x_v)\in(\mathcal{D}_f, \mathcal{D}_v)$
    \STATE Update attacker by stochastic gradient ascent as \\ 
    \vspace{-.15in}
    \begin{equation*}
        \theta_{\mathbf{A}} \gets \theta_{\mathbf{A}} + \eta_{\mathbf{A}} \triangledown_{\mathbf{\theta_{\mathbf{A}}}}\mathcal{L}_{\mathbf{A}}, \text{ with } \mathcal{L}_{\mathbf{A}} = \frac{1}{B}\sum \log(\mathbf{A}(\mathbf{D}(x_f)\prime)+\log(1-\mathbf{A}(\mathbf{D}(x_v)\prime)
    \end{equation*}
    \vspace{-.17in}
    \STATE Update defender by stochastic gradient descent as \\
    \vspace{-.17in}
           \begin{equation*}
               \theta_{\mathbf{D}} \gets \theta_{\mathbf{D}} - \eta_{\mathbf{D}} \triangledown_{\mathbf{\theta_{\mathbf{D}}}}(\mathcal{L}_{\mathbf{D}}+\alpha\mathcal{L}_{\mathbf{A}}+\beta\mathcal{V}_{ss}), \text{ with } \mathcal{L}_{\mathbf{D}} = \frac{1}{B}\sum \Big[-\sum_{i=1}^{K} y_{ri} \cdot \log(\mathbf{D}(x_r)_i)\Big] 
           \end{equation*}
    \vspace{-.12in}
\ENDFOR
\end{algorithmic}
\end{algorithm*}

\begin{table*}[!ht] 
\setlength{\fboxsep}{0pt}
\centering
\fontsize{9}{11}\selectfont
\setlength{\tabcolsep}{3pt} 
\setlength\arrayrulewidth{0.6pt}
\sisetup{table-format=2.2} 
\caption{\textbf{Cifar-10, Cifar-100 comparison.} We evaluate the unlearning performance on Cifar datasets. The top two tables encompass the unlearning performances on Cifar-10 for the methods listed for random and class-wise schemes respectively. Evaluation metrics from section \ref{sec:evaluation} are employed, including the averaged disparity of performances relative to retraining on the retain set $\mathcal{D}r$. Five separate seeds are utilized for each method, and the reported mean and standard deviation are presented in the format $a{\pm b}$. Additionally, unlearning is performed on $95\%$ sparse models for each method, employing OMP \citep{ma2021sanity} to enforce sparsity. The relative performances are compared against the "retrain" standard, and the average disparity provides an overarching assessment of all algorithms across the employed metrics. (\textbf{UA:} Unlearning Acc., \textbf{RA:} Retaining Acc., \textbf{TA:} Test Acc.)}
\vspace{0.05in}
\resizebox{\textwidth}{!}{%
\colorbox{LightGray}{\begin{tabular}{lccccccccccc}
\toprule
 & \multicolumn{11}{c}{Cifar-10 Random Forgetting} \\ 
\multirow{2}{*}{\textbf{Methods}} & \multicolumn{2}{c}{\textbf{UA}} & \multicolumn{2}{c}{\textbf{MIA-Efficacy}} & \multicolumn{2}{c}{\textbf{RA}} & \multicolumn{2}{c}{\textbf{TA}} & \multicolumn{2}{c}{\textbf{Avg. Disparity}} & \multirow{2}{*}{\textbf{Run Time}} \\ \cmidrule{2-11}
 & {Dense} & {95\% Sparse} & {Dense} & {95\% Sparse} & {Dense} & {95\% Sparse} & {Dense} & {95\% Sparse} & {Dense} & {95\% Sparse} & {min.} \\
\midrule
\rowcolor{LightRed}\multicolumn{1}{c|}{Retrain} &$5.80_{\pm0.12}$ & $6.81_{\pm0.23}$ &$13.91_{\pm0.15}$& $15.17_{\pm0.15}$& $100.00_{\pm0.00}$ & $100.00_{\pm0.00}$& $94.30_{\pm0.13}$ &$92.21_{\pm0.22}$ &0.00 &0.00 & 82.15 \\
\rowcolor{LLightGray}\multicolumn{1}{c|}{FT}      & $0.18_{\pm0.04}$ & $0.04_{\pm0.04}$& $1.70_{\pm0.10}$ & $1.95_{\pm0.01}$& $99.92_{\pm0.07}$ & $99.98_{\pm0.01}$& $94.25_{\pm0.15}$ & $94.13_{\pm0.11}$& $4.48$ & $4.52$& 4.91 \\
\multicolumn{1}{c|}{GA}      & \cellcolor{LightGray}$0.00_{\pm0.00}$ &$0.03_{\pm0.01}$ & $0.35_{\pm0.15}$ & $0.69_{0.03}$& $99.99_{\pm0.01}$ &$100.00_{\pm0.00}$ & $94.80_{\pm0.03}$ &$94.40_{\pm0.04}$ & 4.96 & 5.86 & 0.37 \\
\rowcolor{LLightGray}\multicolumn{1}{c|}{FF}      & $6.96_{\pm1.15}$ & $10.11_{\pm0.48}$& $10.24_{\pm0.52}$ &$10.47_{\pm0.48}$ & $93.09_{\pm1.06}$ &$90.02_{\pm0.05}$ & $88.51_{\pm0.98}$ &$85.13_{\pm0.01}$ & 4.37 & 6.26 & 42.9 \\
\multicolumn{1}{c|}{IU}      & \cellcolor{LightGray}$0.42_{\pm0.78}$ &$0.48_{\pm0.36}$ & $1.49_{\pm1.18}$ &$2.58_{\pm0.75}$ & $99.57_{\pm0.81}$ &$99.54_{\pm0.37}$ & $93.89_{\pm1.03}$ &$92.97_{\pm0.55}$ & 4.96 & 5.95&4.93 \\ \hline
\rowcolor{LightCyan}\multicolumn{1}{c|}{Ours}    & $3.46_{\pm0.13}$ &$4.92_{\pm0.26}$ & $8.25_{\pm0.02}$ & $12.27_{\pm0.00}$& $99.50_{\pm0.21}$ & $96.78_{\pm0.28}$& $93.50_{\pm0.12}$ &$90.96_{\pm0.24}$ & $2.32$ & $2.32$ & 7.98 \\
\hline 
\hline
 & \multicolumn{11}{c}{Cifar-10 Class-wise Forgetting} \\ 
\multirow{2}{*}{\textbf{Methods}} & \multicolumn{2}{c}{\textbf{UA}} & \multicolumn{2}{c}{\textbf{MIA-Efficacy}} & \multicolumn{2}{c}{\textbf{RA}} & \multicolumn{2}{c}{\textbf{TA}} & \multicolumn{2}{c}{\textbf{Avg. Disparity}} & \multirow{2}{*}{\textbf{Run Time}} \\ \cmidrule{2-11}
 & {Dense} & {95\% Sparse} & {Dense} & {95\% Sparse} & {Dense} & {95\% Sparse} & {Dense} & {95\% Sparse} & {Dense} & {95\% Sparse} & {min.} \\
\midrule
\rowcolor{LightRed}\multicolumn{1}{c|}{Retrain} & $100.00_{\pm0.00}$ & $100.00_{\pm0.00}$& $100.00_{\pm0.00}$ & $100.00_{\pm0.00}$& $100.00_{\pm0.00}$ & $100.00_{\pm0.00}$& $94.81_{\pm0.09}$ &$91.76_{\pm0.93}$ & 0.00 &0.00 & 82.00 \\
\rowcolor{LLightGray}\multicolumn{1}{c|}{FT}      & $8.36_{\pm3.03}$ &$31.31_{\pm17.01}$ & $40.76_{\pm8.03}$ &$75.29_{\pm16.94}$ & $99.92_{\pm0.03}$ &$99.98_{\pm0.01}$ & $94.41_{\pm0.29}$ &$94.48_{\pm0.26}$ & 37.84 & 24.03 & 4.88 \\
\multicolumn{1}{c|}{GA}      & \cellcolor{LightGray} $93.80_{\pm1.96}$ &$96.69_{\pm1.97}$ & $96.06_{\pm1.87}$ &$97.37_{\pm2.24}$ & $93.76_{\pm0.73}$ &$88.01_{\pm2.16}$ & $87.15_{\pm0.58}$ &$82.29_{\pm1.98}$ & 6.00 &6.84 & 0.39 \\
\rowcolor{LLightGray}\multicolumn{1}{c|}{FF}      & $67.57_{\pm13.05}$ &$92.54_{\pm3.59}$ & $100.00_{\pm0.00}$ &$100.00_{\pm0.0}$ & $100.00_{\pm0.00}$ &$100.00_{\pm0.00}$ & $94.98_{\pm0.21}$&$94.75_{\pm0.29}$ & 8.14 & 3.42 & 42.7 \\
\multicolumn{1}{c|}{IU}      & \cellcolor{LightGray}$64.41_{\pm24.94}$ &$95.60_{\pm6.15}$ & $81.66_{\pm17.64}$ &$98.95_{\pm1.81}$ & $98.88_{\pm1.33}$ &$97.99_{\pm1.47}$ & $92.77_{\pm1.75}$ &$91.47_{\pm1.79}$ & 14.27 & 1.94 &4.40\\ \hline
\rowcolor{LightCyan}\multicolumn{1}{c|}{Ours}    & $100.00_{\pm0.00}$ &$100.00_{\pm0.00}$ &$100.00_{\pm0.00}$ &$100.00_{\pm0.00}$ &$98.90_{\pm0.20}$  &$95.00_{\pm0.43}$ & $93.22_{\pm0.74}$ &$90.47_{\pm0.48}$ & 0.67  & 1.57 & 9.91\\

\bottomrule
\end{tabular}}
}

\vspace{0.15in}


\resizebox{\textwidth}{!}{%
\colorbox{LightGray}{\begin{tabular}{lccccccccccc}
\toprule
 & \multicolumn{11}{c}{Cifar-100 Random Forgetting} \\ 
\multirow{2}{*}{\textbf{Methods}} & \multicolumn{2}{c}{\textbf{UA}} & \multicolumn{2}{c}{\textbf{MIA-Efficacy}} & \multicolumn{2}{c}{\textbf{RA}} & \multicolumn{2}{c}{\textbf{TA}} & \multicolumn{2}{c}{\textbf{Avg. Disparity}} & \multirow{2}{*}{\textbf{Run Time}} \\ \cmidrule{2-11}
 & {Dense} & {95\% Sparse} & {Dense} & {95\% Sparse} & {Dense} & {95\% Sparse} & {Dense} & {95\% Sparse} & {Dense} & {95\% Sparse} & {min.} \\
\midrule
\rowcolor{LightRed}\multicolumn{1}{c|}{Retrain} & $24.75_{\pm0.11}$ & $26.93_{\pm1.04}$& $49.68_{\pm0.35}$ & $44.49_{\pm0.44}$& $99.98_{\pm0.01}$ & $99.57_{\pm0.08}$& $74.57_{\pm0.06}$ &$69.73_{\pm0.15}$ & 0.00 &0.00 & 82.33 \\
\rowcolor{LLightGray}\multicolumn{1}{c|}{FT}      & $0.11_{\pm0.03}$ &$2.57_{\pm0.21}$ & $5.66_{\pm0.47}$ &$12.96_{\pm0.25}$ & $99.97_{\pm0.01}$ & $99.57_{\pm0.04}$& $75.45_{\pm0.17}$ & $73.55_{\pm0.17}$&16.65 &13.02 & 4.91 \\
\multicolumn{1}{c|}{GA}      & \cellcolor{LightGray} $0.04_{\pm0.01}$ &$0.21_{\pm0.05}$ & $2.07_{\pm0.15}$ &$4.26_{\pm0.29}$ & $99.98_{\pm0.00}$ &$99.83_{\pm0.02}$ & $75.38_{\pm0.07}$ &$74.33_{\pm0.08}$ &17.58 &15.52 & 0.41 \\
\rowcolor{LLightGray}\multicolumn{1}{c|}{FF}      & $0.05_{\pm0.01}$ &$0.23_{\pm0.06}$ & $2.04_{\pm0.21}$ &$4.12_{\pm0.15}$ & $99.98_{\pm0.00}$ & $99.84_{\pm0.01}$& $75.48_{\pm0.07}$ &$74.26_{\pm0.05}$ &17.56  &15.57 & 42.8 \\
\multicolumn{1}{c|}{IU}      & \cellcolor{LightGray}$0.11_{\pm0.06}$ & $2.49_{\pm0.91}$& $3.65_{\pm0.70}$ &$7.43_{\pm0.89}$ & $99.94_{\pm0.04}$ &$98.07_{\pm0.81}$ & $74.72_{\pm0.34}$ & $71.83_{\pm0.75}$& 17.35 &15.22&5.25 \\ \hline
\rowcolor{LightCyan}\multicolumn{1}{c|}{Ours}    & $10.50_{\pm0.45}$ & $13.36_{\pm0.97}$ &$30.09_{\pm0.01}$ &$24.65_{\pm0.01}$  &$97.60_{\pm0.45}$  &$95.21_{\pm0.54}$ & $71.25_{\pm0.59}$ &$69.94_{\pm0.01}$ &5.48   &5.43 & 13.3 
\\
\hline
\hline
 & \multicolumn{11}{c}{Cifar-100 Class-wise Forgetting} \\ 
\multirow{2}{*}{\textbf{Methods}} & \multicolumn{2}{c}{\textbf{UA}} & \multicolumn{2}{c}{\textbf{MIA-Efficacy}} & \multicolumn{2}{c}{\textbf{RA}} & \multicolumn{2}{c}{\textbf{TA}} & \multicolumn{2}{c}{\textbf{Avg. Disparity}} & \multirow{2}{*}{\textbf{Run Time}} \\ \cmidrule{2-11}
 & {Dense} & {95\% Sparse} & {Dense} & {95\% Sparse} & {Dense} & {95\% Sparse} & {Dense} & {95\% Sparse} & {Dense} & {95\% Sparse} & {min.} \\
\midrule
\rowcolor{LightRed}\multicolumn{1}{c|}{Retrain} & $100.00_{\pm0.00}$ & $100.00_{\pm0.00}$& $100.00_{\pm0.00}$ & $100.00_{\pm0.00}$& $99.98_{\pm0.01}$ & $96.59_{\pm0.12}$& $73.75_{\pm0.20}$ &$69.49_{\pm0.21}$ & 0.00 &0.00 & 82.22 \\
\rowcolor{LLightGray}\multicolumn{1}{c|}{FT}      & $12.04_{\pm7.1}$ & $59.15_{\pm23.02}$ & $70.22_{\pm17.57}$ &$78.67_{\pm19.74}$ & $99.95_{\pm0.01}$ &$98.43_{\pm0.09}$ & $74.60_{\pm0.30}$ &$72.44_{\pm0.32}$ &29.23 &14.35 & 4.85 \\
\multicolumn{1}{c|}{GA}      & \cellcolor{LightGray} $68.67_{\pm1.16}$ &$84.22_{\pm4.19}$ & $91.87_{\pm6.36}$ & $91.56_{\pm5.69}$& $96.55_{\pm3.02}$ &$89.91_{\pm6.20}$ & $68.40_{\pm3.42}$ & $65.85_{\pm4.13}$&12.06 &8.64 & 0.38 \\
\rowcolor{LLightGray}\multicolumn{1}{c|}{FF}      & $0.05_{\pm0.09}$ &$0.17_{\pm0.26}$ & $1.99_{\pm1.13}$ &$4.27_{\pm3.03}$ & $99.98_{\pm0.00}$ &$99.83_{\pm0.00}$ & $75.43_{\pm0.02}$ & $74.34_{\pm0.00}$&49.07  &46.87 & 42.6 \\
\multicolumn{1}{c|}{IU}      & \cellcolor{LightGray}$20.14_{\pm15.76}$ &$84.09_{\pm21.58}$ & $82.67_{\pm9.40}$ & $95.24_{\pm7.49}$& $99.97_{\pm0.02}$ &$96.42_{\pm2.68}$ & $74.69_{\pm0.46}$ &$70.44_{\pm2.24}$ & 24.67  &4.97&4.93 \\ \hline
\rowcolor{LightCyan}\multicolumn{1}{c|}{Ours}    & $97.20_{\pm2.15}$ &$99.27_{\pm0.76}$ & $99.84_{\pm0.02}$  &$99.93_{\pm0.01}$ &$98.78{\pm0.26}$  &$91.52{\pm1.06}$ & $71.49{\pm0.52}$ &$69.60{\pm0.63}$ & 0.89 & 0.85 & 16.65\\

\bottomrule
\end{tabular}}
}
\end{table*}\label{tab:main}

\section{Experiments} \label{sec:evaluation}
\subsection{Criteria}\label{sec:metrics}
In an ideal scenario for machine unlearning, a perfectly functional system ensures the following set of conditions are met.
\begin{itemize}
    \item The model must be able to generalize to other datasets. We evaluate this condition through the accuracy of the scrubbed defender on the test set $\mathcal{D}_t$ Formally, $\mathbf{TA}=Acc_{\mathcal{D}_t}(\mathbf{D}_{\theta_u})$.
    \item The performance of the unlearned model is retained on the  $\mathcal{D}_r$ which is measured through retain set accuracy. Formally, $\mathbf{RA}=Acc_{\mathcal{D}_r}(\mathbf{D}_{\theta_u})$.
    \item The unlearning performance of the model closely approximates the gold standard of retraining, which is measured on the forget set $\mathcal{D}_f$ using unlearning accuracy defined as $\mathbf{UA}=1-Acc_{\mathcal{D}_f}(\mathbf{D}_{\theta_u})$.
    \item The general capacity of the unlearned model to defend against MIA. For this metric we employ a separate prediction confidence based attack, which utilizes the differences in confidence of the trained network for samples arising from $\mathcal{D}$ vs $\mathcal{D}_t$ or $\mathcal{D}_v$. The confidence based predictor \cite{yeom2018privacy, song2019privacy} is formulated through training a support vector machine on the prediction confidences of the network for samples arising from $\mathcal{D}$ and $\mathcal{D}_t$ and subsequently used to predict the origin of $\mathcal{D}_f$. A successful attack will therefore perform with high MIA-Efficacy on $\mathcal{D}_f$. Formally, for an attacker to fail, it must predict samples arising from $\mathcal{D}_f$ as true negatives and therefore, $\mathbf{MIA \text{-}efficacy} = TN/|\mathcal{D}_f|$ where $TN$ denotes the true negative inferences by the attacker.
    \item The efficiency of the algorithm is evaluated by considering the run-time for one random seed of forgetting, ensuring it is not overly cumbersome compared to the gold standard retraining model and other unlearning methods.
\end{itemize}

Our comparisons are set against popular paradigms of machine unlearning described as follows.
\begin{itemize} \label{sec:baselines}
    \item \textit{Fine-tuning (\textbf{FT})} \cite{warnecke2021machine, golatkar2020eternal}: This method involves finetuning the network only on the retain set $\mathcal{D}_r$ in order to induce amnesia over the network's prior training on $\mathcal{D}_f$ leveraging the network's tendency for catastrophic forgetting, and ensconce the performance over the retain set.
    \item \textit{Gradient-ascent (\textbf{GA})} \cite{graves2021amnesiac, thudi2022unrolling}: Using the negative of the classification loss evaluated on $\mathcal{D}_f$, this method retrains the model in order to explicitly negate the prior training and induce forgetting.
    \item \textit{Fisher Forget (\textbf{FF})} \cite{becker2022evaluating, izzo2021approximate}: FF uses estimates the hessian of the defender network using fisher information matrix, in order to informatively corrupt the  of the pre-trained network's parameters $\theta_0$ using additive gaussian noise in the image space.
    \item \textit{Influence Unlearning (\textbf{IU})} \cite{koh2017understanding}: This method uses estimates of influence functions which evaluate the change in parameters of a pre-trained network due to the presence/absence of samples in $\mathcal{D}$ through "upweighing" \cite{cook1980characterizations} these samples on the network parameters, allowing for the method to manipulate the influence of the loss due to $\mathcal{D}_f$ on the defender.
\end{itemize}

\subsection{Configuration}
Training is conducted on a single Nvidia RTX-A6000 GPU. The defender's base model is the ResNet-18 \cite{he2016deep}, while the attacker network adopts a transformer structure with $3$ multihead self-attention layers and $4$ attention heads, retaining the default configuration from \citet{yuan2022membership}. Pre-training of the defender employs a learning rate of $0.1$ for both Cifar datasets. For random forgetting, the (un-)learning rate is set to $0.01$ with the dense model, while a higher rate of $0.03$ is found to be more suitable for a sparse model in this scenario. In the class-wise unlearning scenario, a rate of $0.02$ is used for both levels of sparsity and datasets. Empirical adjustments are made to the $\alpha$ and $\beta$ parameters, setting them to $0.9$ and $0.001$, respectively.

\paragraph{Training Adjustments}
We find it beneficial to pre-train the attacker network for $1000$ iterations to classify the defender's outputs for $\mathcal{D}$ and $\mathcal{D}_v$, enabling the framework to leverage informative gradients provided by the attacker in the early training stages. During class-wise forgetting, we observe that unlearning the class for both datasets is notably easier than retaining general performance. Consequently, we adjust the $\alpha$ parameter to 0 after the first 30 iterations in this scenario.

\begin{figure*}[ht]
\vskip 0.2in
\begin{center}
\includegraphics[width=1.0\textwidth]{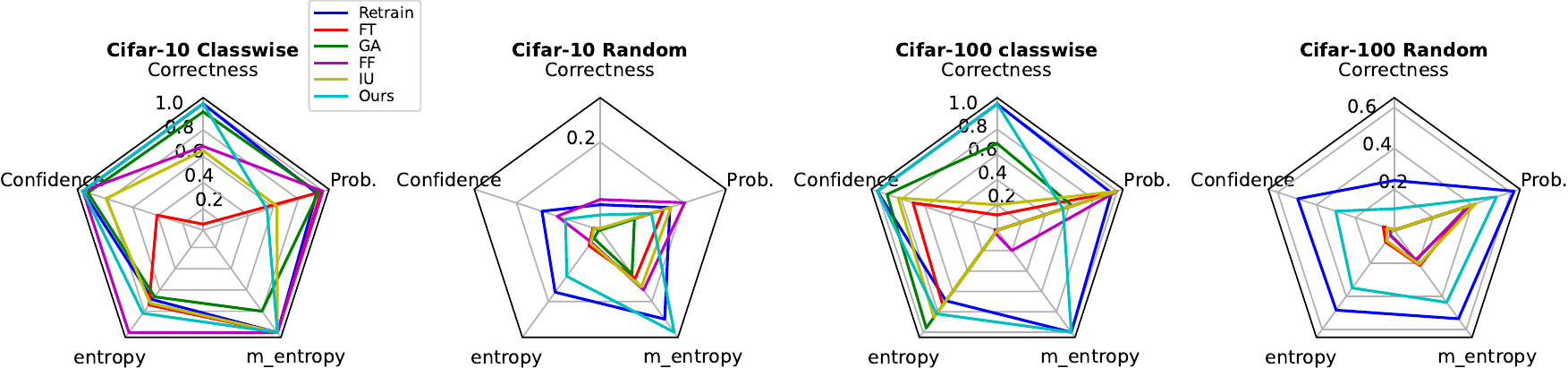}
\caption{\textbf{MIA-Attack Robustness.} We incorporate additional MIA-Attack methods and compare the robustness of our method against the baselines. The attacks are based on Prediction Confidence, Prediction Entropy, Modified Prediction Entropy and Prediction Probability as elucidated in Section \ref{sec:mia_metrics} under Appendix.}
\label{fig:MIA}
\end{center}
\vskip -0.2in
\end{figure*}

\subsection{Results}
Our results are listed in the Table \ref{tab:main} for all the aforementioned configurations. We conduct our evaluations under the random forgetting scheme where $10\%$ of the samples are chosen randomly to be scrubbed from the pre-trained model, and class-wise forgetting scheme where all samples from a randomly chosen class are tasked for removal. All methods were evaluated using the dense model and a $95\%$ sparse model, in order to gauge the robustness and efficacy of the listed methods under varying conditions of sparsity. Sparsity in the networks is enforced using One-Shot Magnitude Pruning (OMP) \cite{ma2021sanity} due to its lower computational overhead and better generalization \cite{jia2023model}. The gold-standard for unlearning which is to retrain the model on $\mathcal{D}_r$ is also evaluated. The assessment employs the metrics outlined in section \ref{sec:metrics}. It is imperative to emphasize that elevated values across these metrics do not inherently denote superior performance and unlearning methods should exhibit a minimal disparity in comparison to the retrain method. Consequently, our evaluation incorporates the computation of the average disparity of scores, aiming to simulate a measure of proximity with this gold standard. The experiments reveal that our proposed method surpasses all baseline approaches across the utilized metrics and attains the utmost proximity to the gold standard. In the context of the random forgetting scheme, our method consistently achieves the closest performance to the retrain method. Simultaneously, it maintains superior performance in both the retain set (\textbf{RA}) and test set (\textbf{TA}). Contrastingly, other baseline methods exhibit ineffectiveness in the task of forgetting, resulting in excessively high scores for \textbf{RA} and \textbf{TA}, or they adopt over-compensatory measures, causing a decline in the model's overall classification capacity,, as evidenced by the \textbf{MIA-efficacy} metric. Notably, FF achieves significant scores in Cifar-10 under this scheme but experiences a considerable decline with over $5\%$ point gap from the retrain benchmark in the \textbf{TA} and \textbf{RA} metrics.

Nevertheless, notable observations arise under the class-wise scheme for both datasets, where the methods demonstrate competitive performances. In the case of the Cifar-10 dataset, a high \textbf{UA} metric is observed across the methods, accompanied by elevated values for \textbf{RA} and \textbf{TA}. Notably, our method achieves flawless performance across the \textbf{UA} and \textbf{MIA-Efficacy} metrics, maintaining a remarkably narrow margin for the \textbf{RA} and \textbf{TA} metrics with merely a $0.67$ score in the \textbf{Avg. Disparity} measure. This achievement mitigates the necessity for sparsification measures to improve performance. For Cifar-100, a comparable performance is also observed across these metrics, while concurrently sustaining a minimal performance gap with the gold standard.

Sparse models however, ease the process of unlearning performance for the methods as noted by \citet{jia2023model} whilst incurring a marginal cost over the \textbf{TA} and \textbf{RA} metrics. This phenomenon is particularly pronounced in class-wise forgetting schemes, where the baseline methods demonstrate significant improvements across the metrics and resulting in substantially lower \textbf{Avg. Disparity} scores. Our approach demonstrates a comparable trend, closely aligning with the retraining method across various metrics. Notably, we argue that incorporating sparsity as a performance enhancement measure yields marginal benefits in our context. The algorithm consistently achieves robust performance with dense variants, presenting challenges for the baselines, even when employing $95\%$ sparse models. Moreover, $\mathcal{V}_{ss}$ plays a critical role in enhancing performance as we elucidate in Section \ref{sec:ssr_ablations}.

\begin{table*}[!ht] 
\setlength{\fboxsep}{0pt}
\centering
\fontsize{9}{11}\selectfont
\setlength{\tabcolsep}{3pt} 
\setlength\arrayrulewidth{0.6pt}
\sisetup{table-format=2.2} 
\caption{\textbf{Ablations for $\mathcal{V}_{ss}$} Here we evaluate our method on the random forgetting scenario over both Cifar-10 and Cifar-100 datasets to illustrate the role of the self-supervised regularization objective in our algorithm. The format follows from Table \ref{tab:main}. (\textbf{UA:} Unlearning Acc., \textbf{RA:} Retaining Acc., \textbf{TA:} Test Acc.)}
\vspace{0.05in}

\resizebox{\textwidth}{!}{%
\colorbox{LightGray}{\begin{tabular}{lccccccccccc}
\toprule
 & \multicolumn{11}{c}{Cifar-10 Random Forgetting} \\ 
\multirow{2}{*}{\textbf{Methods}} & \multicolumn{2}{c}{\textbf{UA}} & \multicolumn{2}{c}{\textbf{MIA-Efficacy}} & \multicolumn{2}{c}{\textbf{RA}} & \multicolumn{2}{c}{\textbf{TA}} & \multicolumn{2}{c}{\textbf{Avg. Disparity}} & \multirow{2}{*}{\textbf{Run Time}} \\ \cmidrule{2-11}
 & {Dense} & {95\% Sparse} & {Dense} & {95\% Sparse} & {Dense} & {95\% Sparse} & {Dense} & {95\% Sparse} & {Dense} & {95\% Sparse} & {min.} \\
\midrule
\rowcolor{LightRed}\multicolumn{1}{c|}{Retrain} &$5.80_{\pm0.12}$ & $6.81_{\pm0.23}$ &$13.91_{\pm0.15}$& $15.17_{\pm0.15}$& $100.00_{\pm0.00}$ & $100.00_{\pm0.00}$& $94.30_{\pm0.13}$ &$92.21_{\pm0.22}$ &0.00 &0.00 & 82.15 \\ \hline
\rowcolor{LightCyan}\multicolumn{1}{c|}{Ours}    & $3.46_{\pm0.13}$ &$4.92_{\pm0.26}$ & $8.25_{\pm0.02}$ & $12.27_{\pm0.01}$& $99.50_{\pm0.21}$ & $96.78_{\pm0.28}$& $93.50_{\pm0.12}$ &$90.96_{\pm0.24}$ & $2.32$ & $2.32$ & 7.98 \\
\rowcolor{LightCyan}\multicolumn{1}{c|}{Ours (No $\mathcal{V}_{ss}$)}    & $3.03_{\pm0.23}$ &$4.13_{\pm0.28}$ & $7.60_{\pm0.02}$ & $11.67_{\pm0.02}$& $99.64_{\pm0.21}$ & $97.04_{\pm0.28}$& $93.54_{\pm0.19}$ &$91.04_{\pm0.45}$ & $2.61$ & $2.59$ & 7.83 \\
\hline 
\hline
 & \multicolumn{11}{c}{Cifar-100 Random Forgetting} \\ 
\multirow{2}{*}{\textbf{Methods}} & \multicolumn{2}{c}{\textbf{UA}} & \multicolumn{2}{c}{\textbf{MIA-Efficacy}} & \multicolumn{2}{c}{\textbf{RA}} & \multicolumn{2}{c}{\textbf{TA}} & \multicolumn{2}{c}{\textbf{Avg. Disparity}} & \multirow{2}{*}{\textbf{Run Time}} \\ \cmidrule{2-11}
 & {Dense} & {95\% Sparse} & {Dense} & {95\% Sparse} & {Dense} & {95\% Sparse} & {Dense} & {95\% Sparse} & {Dense} & {95\% Sparse} & {min.} \\
\midrule
\rowcolor{LightRed}\multicolumn{1}{c|}{Retrain} & $24.75_{\pm0.11}$ & $26.93_{\pm1.04}$& $49.68_{\pm0.35}$ & $44.49_{\pm0.44}$& $99.98_{\pm0.01}$ & $99.57_{\pm0.08}$& $74.57_{\pm0.06}$ &$69.73_{\pm0.15}$ & 0.00 &0.00 & 82.33 \\\hline
\rowcolor{LightCyan}\multicolumn{1}{c|}{Ours}    & $10.50_{\pm0.45}$ & $13.36_{\pm0.97}$ &$30.09_{\pm0.01}$ &$24.65_{\pm0.01}$  &$97.60_{\pm0.45}$  &$95.21_{\pm0.54}$ & $71.25_{\pm0.59}$ &$69.94_{\pm0.01}$ &5.48   &5.43 & 13.3 \\
\rowcolor{LightCyan}\multicolumn{1}{c|}{Ours (No $\mathcal{V}_{ss}$)}    & $9.79_{\pm0.51}$ & $13.34_{\pm0.97}$ &$29.10_{\pm0.32}$ &$23.34_{\pm0.07}$  &$97.62_{\pm0.41}$  &$95.73_{\pm0.53}$ & $71.28_{\pm0.44}$ &$70.62_{\pm0.05}$ &5.71   &5.57 & 12.3 
\\
\bottomrule
\end{tabular}}
}
\end{table*}\label{tab:ablations}

\subsection{MIA-Robustness} \label{sec:mia_metrics}
We perform Membership Inference Attacks (MIAs) based on metrics on the resulting unlearned model. These attacks involve deducing membership by calculating various metrics on the prediction vectors. Unlike binary classification predictions based on outputs of the defender network, metric based MIA leverage informative metrics in order to elucidate intricate patterns in the defender's outputs for members and non members. We incorporate four distinct metrics, described as follows.
\begin{itemize}
    \item Prediction correctness MIA \cite{yeom2018privacy}: The attacker is trained to infer membership of a record if the defender's predictions are correct. This attack capitalizes on the defender's ability to predict accurately on $\mathcal{D}$ which may not generalize to $\mathcal{D}_t$.
    \item Prediction confidence MIA \cite{salem2018mlleaks}: The attacker infers membership based on the maximum prediction confidence scores of the defender across all samples. This leverages differential scores of confidence measures for the defender over $\mathcal{D}$ and $\mathcal{D}_t$.
    \item Prediction entropy MIA \cite{salem2018mlleaks}: The attacker evaluates the entropy of predictions over samples from $\mathcal{D}$ and $\mathcal{D}_t$, considering the possibility that the entropy of predictions for $\mathcal{D}_t$ is higher.
    \item Modified prediction entropy MIA \cite{song2019privacy}: In addition to prediction entropy, the attacker incorporates ground truth label information to leverage the defender's capacity for false negatives, inferring membership based on, for example, incorrectly predicted records.
\end{itemize}

The attackers are trained separately using SVMs over inferences from $\mathbf{D}_{\theta_u}$, The illustration is provided in Figure \ref{fig:MIA}. We notice that the framework having received explicit information solely in the form of prediction scores and ground truth labels manages to generalize to foreign attackers trained independently, consistently exhibiting superior performances across the metrics and yet again closely approximating the retrained model.

 \subsection{Role of \texorpdfstring{$\mathcal{V}_{ss}$}{Lg}}\label{sec:ssr_ablations}
 We conduct ablations in order to evaluate the role of the self supervised objective (Section \ref{sec:ssr}) towards the overall goal of effective machine unlearning and robust generalization. The ablations are listed in Table \ref{tab:ablations}. We notice that the incorporation of the objective consistently yields a higher \textbf{UA} and \textbf{MIA-efficacy}, with marginally lower scores under the \textbf{RA} and \textbf{TA} metrics. Henceforth, this objective serves as an optional augmentation lending to overall benefits as measured by \textbf{Avg. Disparity} at the price of higher run-times. Nevertheless, the performance of the method remains robust and one may opt to incorporate the objective within the framework through marginal adjustments using the parameter $\beta$, as per the use case and empirical needs.

\section{Conclusion}
In this study, we employ a straightforward and differentiable framework, adhering to the well-established principles of min-max optimization, to address the challenge of machine unlearning. This framework demonstrates high efficiency, yielding substantial runtime improvements over retraining (at least 5x in the case of Cifar-100 class-wise). Moreover, its effectiveness is evident through consistently achieving the lowest disparity scores across various configurations. Remarkably, the proposed method attains near-perfect performance, particularly notable in the class-wise forgetting schemes. In these scenarios, the unlearning accuracy and MIA robustness reach 100\% for Cifar-10 and approach this level for Cifar-100, even within dense network regimes with minimal compromises observed in the test accuracies. Furthermore, the method avoids the expenses of hessian matrix approximations for deep networks and involves minimal parameter tuning requirements. The approach additionally, is adaptable to incorporation of advanced MIA attack techniques and improvements over the objectives of the defender and attacker in order to ensure better performances and convergence. Nevertheless, it is noteworthy that the framework excels under the class-wise unlearning scheme, leaving room for refinement in the context of the random unlearning scheme—potentially a more realistic scenario for machine unlearning. We hypothesize that integrating improved optimization procedures \cite{arjovsky2017wasserstein, gulrajani2017improved, kodali2017convergence} and employing stronger, multiple attackers leveraging additional information from predictions \cite{yeom2018privacy, salem2018mlleaks, song2021systematic} could enhance the robustness of the framework in this scheme. This avenue remains a subject for future exploration.

\section{Broader Impact}
This paper on machine unlearning introduces impactful advancements with far-reaching implications. Our proposed techniques not only enhance privacy protection by enabling the removal or modification of sensitive information from trained models but also contribute to fostering fairness by allowing models to adapt and rectify biases over time. Moreover, the efficiency of our machine unlearning methods, in contrast to traditional retraining approaches, aligns with sustainability goals by reducing computational resource requirements. The transparent and accountable nature of our unlearning techniques enhances trust in AI technologies, crucial for widespread acceptance and cooperation between AI systems and users. As researchers and practitioners, we emphasize the responsible navigation of these impacts, ensuring that our machine unlearning strategies adhere to ethical principles and societal values.

\bibliography{icml2024}

\begin{thebibliography}{63}
\providecommand{\natexlab}[1]{#1}
\providecommand{\url}[1]{\texttt{#1}}
\expandafter\ifx\csname urlstyle\endcsname\relax
  \providecommand{\doi}[1]{doi: #1}\else
  \providecommand{\doi}{doi: \begingroup \urlstyle{rm}\Url}\fi

\bibitem[Al-Harrasi et~al.(2023)Al-Harrasi, Shaikh, and Al-Badi]{al2023towards}
Abir Al-Harrasi, Abdul~Khalique Shaikh, and Ali Al-Badi.
\newblock Towards protecting organisations’ data by preventing data theft by malicious insiders.
\newblock \emph{International Journal of Organizational Analysis}, 31\penalty0 (3):\penalty0 875--888, 2023.

\bibitem[Arjovsky et~al.(2017)Arjovsky, Chintala, and Bottou]{arjovsky2017wasserstein}
Martin Arjovsky, Soumith Chintala, and L{\'e}on Bottou.
\newblock Wasserstein generative adversarial networks.
\newblock In \emph{International conference on machine learning}, pages 214--223. PMLR, 2017.

\bibitem[Bardes et~al.(2021)Bardes, Ponce, and LeCun]{DBLP:journals/corr/abs-2105-04906}
Adrien Bardes, Jean Ponce, and Yann LeCun.
\newblock Vicreg: Variance-invariance-covariance regularization for self-supervised learning.
\newblock \emph{CoRR}, abs/2105.04906, 2021.
\newblock URL \url{https://arxiv.org/abs/2105.04906}.

\bibitem[Becker and Liebig(2022)]{becker2022evaluating}
Alexander Becker and Thomas Liebig.
\newblock Evaluating machine unlearning via epistemic uncertainty.
\newblock \emph{arXiv preprint arXiv:2208.10836}, 2022.

\bibitem[Bourtoule et~al.(2021)Bourtoule, Chandrasekaran, Choquette-Choo, Jia, Travers, Zhang, Lie, and Papernot]{bourtoule2021machine}
Lucas Bourtoule, Varun Chandrasekaran, Christopher~A Choquette-Choo, Hengrui Jia, Adelin Travers, Baiwu Zhang, David Lie, and Nicolas Papernot.
\newblock Machine unlearning.
\newblock In \emph{2021 IEEE Symposium on Security and Privacy (SP)}, pages 141--159. IEEE, 2021.

\bibitem[Cao and Yang(2015)]{7163042}
Yinzhi Cao and Junfeng Yang.
\newblock Towards making systems forget with machine unlearning.
\newblock In \emph{2015 IEEE Symposium on Security and Privacy}, pages 463--480, 2015.
\newblock \doi{10.1109/SP.2015.35}.

\bibitem[Caron et~al.(2021)Caron, Touvron, Misra, J\'egou, Mairal, Bojanowski, and Joulin]{Caron_2021_ICCV}
Mathilde Caron, Hugo Touvron, Ishan Misra, Herv\'e J\'egou, Julien Mairal, Piotr Bojanowski, and Armand Joulin.
\newblock Emerging properties in self-supervised vision transformers.
\newblock In \emph{Proceedings of the IEEE/CVF International Conference on Computer Vision (ICCV)}, pages 9650--9660, October 2021.

\bibitem[Carpentier et~al.(2021)Carpentier, Popa, and Anciaux]{carpentier2021poster}
Robin Carpentier, Iulian~Sandu Popa, and Nicolas Anciaux.
\newblock Poster: Reducing data leakage on personal data management systems.
\newblock In \emph{2021 IEEE European Symposium on Security and Privacy (EuroS\&P)}, pages 716--718. IEEE, 2021.

\bibitem[Che et~al.(2023)Che, Zhou, Zhang, Lyu, Liu, Yan, Dou, and Huan]{che2023fast}
Tianshi Che, Yang Zhou, Zijie Zhang, Lingjuan Lyu, Ji~Liu, Da~Yan, Dejing Dou, and Jun Huan.
\newblock Fast federated machine unlearning with nonlinear functional theory.
\newblock In \emph{International conference on machine learning}, pages 4241--4268. PMLR, 2023.

\bibitem[Chen et~al.(2023)Chen, Gao, Liu, Peng, and Wang]{chen2023boundary}
Min Chen, Weizhuo Gao, Gaoyang Liu, Kai Peng, and Chen Wang.
\newblock Boundary unlearning: Rapid forgetting of deep networks via shifting the decision boundary.
\newblock In \emph{Proceedings of the IEEE/CVF Conference on Computer Vision and Pattern Recognition}, pages 7766--7775, 2023.

\bibitem[Chen et~al.(2020)Chen, Kornblith, Norouzi, and Hinton]{https://doi.org/10.48550/arxiv.2002.05709}
Ting Chen, Simon Kornblith, Mohammad Norouzi, and Geoffrey Hinton.
\newblock A simple framework for contrastive learning of visual representations, 2020.
\newblock URL \url{https://arxiv.org/abs/2002.05709}.

\bibitem[Cook and Weisberg(1980)]{cook1980characterizations}
R~Dennis Cook and Sanford Weisberg.
\newblock Characterizations of an empirical influence function for detecting influential cases in regression.
\newblock \emph{Technometrics}, 22\penalty0 (4):\penalty0 495--508, 1980.

\bibitem[Dai et~al.(2021)Dai, Dong, Hao, Sui, and Wei]{DBLP:journals/corr/abs-2104-08696}
Damai Dai, Li~Dong, Yaru Hao, Zhifang Sui, and Furu Wei.
\newblock Knowledge neurons in pretrained transformers.
\newblock \emph{CoRR}, abs/2104.08696, 2021.
\newblock URL \url{https://arxiv.org/abs/2104.08696}.

\bibitem[Dang(2021)]{dang2021right}
Quang-Vinh Dang.
\newblock Right to be forgotten in the age of machine learning.
\newblock In \emph{Advances in Digital Science: ICADS 2021}, pages 403--411. Springer, 2021.

\bibitem[Dwibedi et~al.(2021)Dwibedi, Aytar, Tompson, Sermanet, and Zisserman]{dwibedi2021little}
Debidatta Dwibedi, Yusuf Aytar, Jonathan Tompson, Pierre Sermanet, and Andrew Zisserman.
\newblock With a little help from my friends: Nearest-neighbor contrastive learning of visual representations.
\newblock In \emph{Proceedings of the IEEE/CVF International Conference on Computer Vision}, pages 9588--9597, 2021.

\bibitem[{European Parliament} and {Council of the European Union}(2016)]{EuropeanParliament2016a}
{European Parliament} and {Council of the European Union}.
\newblock Regulation ({EU}) 2016/679 of the {European} {Parliament} and of the {Council}, 2016.
\newblock URL \url{https://data.europa.eu/eli/reg/2016/679/oj}.

\bibitem[Fredrikson et~al.(2015)Fredrikson, Jha, and Ristenpart]{fredrikson2015model}
Matt Fredrikson, Somesh Jha, and Thomas Ristenpart.
\newblock Model inversion attacks that exploit confidence information and basic countermeasures.
\newblock In \emph{Proceedings of the 22nd ACM SIGSAC conference on computer and communications security}, pages 1322--1333, 2015.

\bibitem[Gandikota et~al.(2023)Gandikota, Materzynska, Fiotto-Kaufman, and Bau]{gandikota2023erasing}
Rohit Gandikota, Joanna Materzynska, Jaden Fiotto-Kaufman, and David Bau.
\newblock Erasing concepts from diffusion models.
\newblock \emph{arXiv e-prints}, pages arXiv--2303, 2023.

\bibitem[Ganju et~al.(2018)Ganju, Wang, Yang, Gunter, and Borisov]{ganju2018property}
Karan Ganju, Qi~Wang, Wei Yang, Carl~A Gunter, and Nikita Borisov.
\newblock Property inference attacks on fully connected neural networks using permutation invariant representations.
\newblock In \emph{Proceedings of the 2018 ACM SIGSAC conference on computer and communications security}, pages 619--633, 2018.

\bibitem[Garc{\'\i}a-Pablos et~al.(2020)Garc{\'\i}a-Pablos, Perez, and Cuadros]{garcia2020sensitive}
Aitor Garc{\'\i}a-Pablos, Naiara Perez, and Montse Cuadros.
\newblock Sensitive data detection and classification in spanish clinical text: Experiments with bert.
\newblock \emph{arXiv preprint arXiv:2003.03106}, 2020.

\bibitem[Geva et~al.(2020)Geva, Schuster, Berant, and Levy]{DBLP:journals/corr/abs-2012-14913}
Mor Geva, Roei Schuster, Jonathan Berant, and Omer Levy.
\newblock Transformer feed-forward layers are key-value memories.
\newblock \emph{CoRR}, abs/2012.14913, 2020.
\newblock URL \url{https://arxiv.org/abs/2012.14913}.

\bibitem[Ginart et~al.(2019)Ginart, Guan, Valiant, and Zou]{NEURIPS2019_cb79f8fa}
Antonio Ginart, Melody Guan, Gregory Valiant, and James~Y Zou.
\newblock Making ai forget you: Data deletion in machine learning.
\newblock In H.~Wallach, H.~Larochelle, A.~Beygelzimer, F.~d\textquotesingle Alch\'{e}-Buc, E.~Fox, and R.~Garnett, editors, \emph{Advances in Neural Information Processing Systems}, volume~32. Curran Associates, Inc., 2019.
\newblock URL \url{https://proceedings.neurips.cc/paper_files/paper/2019/file/cb79f8fa58b91d3af6c9c991f63962d3-Paper.pdf}.

\bibitem[Golatkar et~al.(2020)Golatkar, Achille, and Soatto]{golatkar2020eternal}
Aditya Golatkar, Alessandro Achille, and Stefano Soatto.
\newblock Eternal sunshine of the spotless net: Selective forgetting in deep networks.
\newblock In \emph{Proceedings of the IEEE/CVF Conference on Computer Vision and Pattern Recognition}, pages 9304--9312, 2020.

\bibitem[Goodfellow et~al.(2020)Goodfellow, Pouget-Abadie, Mirza, Xu, Warde-Farley, Ozair, Courville, and Bengio]{goodfellow2020generative}
Ian Goodfellow, Jean Pouget-Abadie, Mehdi Mirza, Bing Xu, David Warde-Farley, Sherjil Ozair, Aaron Courville, and Yoshua Bengio.
\newblock Generative adversarial networks.
\newblock \emph{Communications of the ACM}, 63\penalty0 (11):\penalty0 139--144, 2020.

\bibitem[Graves et~al.(2021)Graves, Nagisetty, and Ganesh]{graves2021amnesiac}
Laura Graves, Vineel Nagisetty, and Vijay Ganesh.
\newblock Amnesiac machine learning.
\newblock In \emph{Proceedings of the AAAI Conference on Artificial Intelligence}, volume~35, pages 11516--11524, 2021.

\bibitem[Gulrajani et~al.(2017)Gulrajani, Ahmed, Arjovsky, Dumoulin, and Courville]{gulrajani2017improved}
Ishaan Gulrajani, Faruk Ahmed, Martin Arjovsky, Vincent Dumoulin, and Aaron~C Courville.
\newblock Improved training of wasserstein gans.
\newblock \emph{Advances in neural information processing systems}, 30, 2017.

\bibitem[He et~al.(2016)He, Zhang, Ren, and Sun]{he2016deep}
Kaiming He, Xiangyu Zhang, Shaoqing Ren, and Jian Sun.
\newblock Deep residual learning for image recognition.
\newblock In \emph{Proceedings of the IEEE conference on computer vision and pattern recognition}, pages 770--778, 2016.

\bibitem[Heng and Soh(2023{\natexlab{a}})]{heng2023continual}
Alvin Heng and Harold Soh.
\newblock Continual learning for forgetting in deep generative models.
\newblock \emph{arXiv preprint}, 2023{\natexlab{a}}.

\bibitem[Heng and Soh(2023{\natexlab{b}})]{heng2023selective}
Alvin Heng and Harold Soh.
\newblock Selective amnesia: A continual learning approach to forgetting in deep generative models.
\newblock \emph{arXiv preprint arXiv:2305.10120}, 2023{\natexlab{b}}.

\bibitem[Hoang et~al.(2024)Hoang, Rana, Gupta, and Venkatesh]{Hoang_2024_WACV}
Tuan Hoang, Santu Rana, Sunil Gupta, and Svetha Venkatesh.
\newblock Learn to unlearn for deep neural networks: Minimizing unlearning interference with gradient projection.
\newblock In \emph{Proceedings of the IEEE/CVF Winter Conference on Applications of Computer Vision (WACV)}, pages 4819--4828, January 2024.

\bibitem[Hoofnagle et~al.(2019)Hoofnagle, Van Der~Sloot, and Borgesius]{hoofnagle2019european}
Chris~Jay Hoofnagle, Bart Van Der~Sloot, and Frederik~Zuiderveen Borgesius.
\newblock The european union general data protection regulation: what it is and what it means.
\newblock \emph{Information \& Communications Technology Law}, 28\penalty0 (1):\penalty0 65--98, 2019.

\bibitem[Izzo et~al.(2021)Izzo, Smart, Chaudhuri, and Zou]{izzo2021approximate}
Zachary Izzo, Mary~Anne Smart, Kamalika Chaudhuri, and James Zou.
\newblock Approximate data deletion from machine learning models.
\newblock In \emph{International Conference on Artificial Intelligence and Statistics}, pages 2008--2016. PMLR, 2021.

\bibitem[Jia et~al.(2023)Jia, Liu, Ram, Yao, Liu, Liu, Sharma, and Liu]{jia2023model}
Jinghan Jia, Jiancheng Liu, Parikshit Ram, Yuguang Yao, Gaowen Liu, Yang Liu, Pranay Sharma, and Sijia Liu.
\newblock Model sparsity can simplify machine unlearning.
\newblock In \emph{Thirty-seventh Conference on Neural Information Processing Systems}, 2023.
\newblock URL \url{https://openreview.net/forum?id=0jZH883i34}.

\bibitem[Kodali et~al.(2017)Kodali, Abernethy, Hays, and Kira]{kodali2017convergence}
Naveen Kodali, Jacob Abernethy, James Hays, and Zsolt Kira.
\newblock On convergence and stability of gans.
\newblock \emph{arXiv preprint arXiv:1705.07215}, 2017.

\bibitem[Koh and Liang(2017)]{koh2017understanding}
Pang~Wei Koh and Percy Liang.
\newblock Understanding black-box predictions via influence functions.
\newblock In \emph{International conference on machine learning}, pages 1885--1894. PMLR, 2017.

\bibitem[Leino and Fredrikson(2020)]{255348}
Klas Leino and Matt Fredrikson.
\newblock Stolen memories: Leveraging model memorization for calibrated {White-Box} membership inference.
\newblock In \emph{29th USENIX Security Symposium (USENIX Security 20)}, pages 1605--1622. USENIX Association, August 2020.
\newblock ISBN 978-1-939133-17-5.
\newblock URL \url{https://www.usenix.org/conference/usenixsecurity20/presentation/leino}.

\bibitem[Lin et~al.(2023)Lin, Zhang, Chen, Chen, and Susilo]{lin2023erm}
Shen Lin, Xiaoyu Zhang, Chenyang Chen, Xiaofeng Chen, and Willy Susilo.
\newblock Erm-ktp: Knowledge-level machine unlearning via knowledge transfer.
\newblock In \emph{Proceedings of the IEEE/CVF Conference on Computer Vision and Pattern Recognition}, pages 20147--20155, 2023.

\bibitem[Liu et~al.(2023{\natexlab{a}})Liu, Xue, Lou, Zhang, Xiong, and Qin]{liu2023muter}
Junxu Liu, Mingsheng Xue, Jian Lou, Xiaoyu Zhang, Li~Xiong, and Zhan Qin.
\newblock Muter: Machine unlearning on adversarially trained models.
\newblock In \emph{Proceedings of the IEEE/CVF International Conference on Computer Vision}, pages 4892--4902, 2023{\natexlab{a}}.

\bibitem[Liu et~al.(2023{\natexlab{b}})Liu, Sun, Wu, and Zhou]{liu2023unlearning}
Yufang Liu, Changzhi Sun, Yuanbin Wu, and Aimin Zhou.
\newblock Unlearning with fisher masking, 2023{\natexlab{b}}.

\bibitem[Liu et~al.(2017)Liu, Tang, Wang, and Chen]{liu2017identification}
Zengjian Liu, Buzhou Tang, Xiaolong Wang, and Qingcai Chen.
\newblock De-identification of clinical notes via recurrent neural network and conditional random field.
\newblock \emph{Journal of biomedical informatics}, 75:\penalty0 S34--S42, 2017.

\bibitem[Ma et~al.(2021)Ma, Yuan, Shen, Chen, Chen, Chen, Liu, Qin, Liu, Wang, et~al.]{ma2021sanity}
Xiaolong Ma, Geng Yuan, Xuan Shen, Tianlong Chen, Xuxi Chen, Xiaohan Chen, Ning Liu, Minghai Qin, Sijia Liu, Zhangyang Wang, et~al.
\newblock Sanity checks for lottery tickets: Does your winning ticket really win the jackpot?
\newblock \emph{Advances in Neural Information Processing Systems}, 34:\penalty0 12749--12760, 2021.

\bibitem[Mehta et~al.(2022)Mehta, Pal, Singh, and Ravi]{mehta2022deep}
Ronak Mehta, Sourav Pal, Vikas Singh, and Sathya~N Ravi.
\newblock Deep unlearning via randomized conditionally independent hessians.
\newblock In \emph{Proceedings of the IEEE/CVF Conference on Computer Vision and Pattern Recognition}, pages 10422--10431, 2022.

\bibitem[Meng et~al.(2022)Meng, Bau, Andonian, and Belinkov]{NEURIPS2022_6f1d43d5}
Kevin Meng, David Bau, Alex Andonian, and Yonatan Belinkov.
\newblock Locating and editing factual associations in gpt.
\newblock In S.~Koyejo, S.~Mohamed, A.~Agarwal, D.~Belgrave, K.~Cho, and A.~Oh, editors, \emph{Advances in Neural Information Processing Systems}, volume~35, pages 17359--17372. Curran Associates, Inc., 2022.
\newblock URL \url{https://proceedings.neurips.cc/paper_files/paper/2022/file/6f1d43d5a82a37e89b0665b33bf3a182-Paper-Conference.pdf}.

\bibitem[Nasr et~al.(2018)Nasr, Shokri, and Houmansadr]{nasr2018machine}
Milad Nasr, Reza Shokri, and Amir Houmansadr.
\newblock Machine learning with membership privacy using adversarial regularization.
\newblock In \emph{Proceedings of the 2018 ACM SIGSAC conference on computer and communications security}, pages 634--646, 2018.

\bibitem[Nguy{\^e}n(2019)]{nguyen2019debunking}
Th{\`a}nh~T{\^a}m Nguy{\^e}n.
\newblock Debunking misinformation on the web: Detection, validation, and visualisation.
\newblock Technical report, EPFL, 2019.

\bibitem[Nguyen et~al.(2021)Nguyen, Nguyen, Nguyen, Vo, Jo, and Nguyen]{nguyen2021judo}
Thanh~Toan Nguyen, Thanh~Tam Nguyen, Thanh~Thi Nguyen, Bay Vo, Jun Jo, and Quoc Viet~Hung Nguyen.
\newblock Judo: Just-in-time rumour detection in streaming social platforms.
\newblock \emph{Information Sciences}, 570:\penalty0 70--93, 2021.

\bibitem[Rosen(2011)]{rosen2011right}
Jeffrey Rosen.
\newblock The right to be forgotten.
\newblock \emph{Stan. L. Rev. Online}, 64:\penalty0 88, 2011.

\bibitem[Salem et~al.(2018)Salem, Zhang, Humbert, Berrang, Fritz, and Backes]{salem2018mlleaks}
Ahmed Salem, Yang Zhang, Mathias Humbert, Pascal Berrang, Mario Fritz, and Michael Backes.
\newblock Ml-leaks: Model and data independent membership inference attacks and defenses on machine learning models, 2018.

\bibitem[Sharma et~al.(2024)Sharma, Ji, Xu, and Chen]{anonymous2024auccl}
Rohan Sharma, Kaiyi Ji, Zhiqiang Xu, and Changyou Chen.
\newblock {AUC}-{CL}: A batchsize-robust framework for self-supervised contrastive representation learning.
\newblock In \emph{The Twelfth International Conference on Learning Representations}, 2024.
\newblock URL \url{https://openreview.net/forum?id=YgMdDQB09U}.

\bibitem[Shokri et~al.(2017)Shokri, Stronati, Song, and Shmatikov]{7958568}
Reza Shokri, Marco Stronati, Congzheng Song, and Vitaly Shmatikov.
\newblock Membership inference attacks against machine learning models.
\newblock In \emph{2017 IEEE Symposium on Security and Privacy (SP)}, pages 3--18, 2017.
\newblock \doi{10.1109/SP.2017.41}.

\bibitem[Song and Mittal(2021)]{song2021systematic}
Liwei Song and Prateek Mittal.
\newblock Systematic evaluation of privacy risks of machine learning models.
\newblock In \emph{30th USENIX Security Symposium (USENIX Security 21)}, pages 2615--2632, 2021.

\bibitem[Song et~al.(2019)Song, Shokri, and Mittal]{song2019privacy}
Liwei Song, Reza Shokri, and Prateek Mittal.
\newblock Privacy risks of securing machine learning models against adversarial examples.
\newblock In \emph{Proceedings of the 2019 ACM SIGSAC Conference on Computer and Communications Security}, pages 241--257, 2019.

\bibitem[Tankard(2016)]{tankard2016gdpr}
Colin Tankard.
\newblock What the gdpr means for businesses.
\newblock \emph{Network Security}, 2016\penalty0 (6):\penalty0 5--8, 2016.

\bibitem[Thudi et~al.(2022)Thudi, Deza, Chandrasekaran, and Papernot]{thudi2022unrolling}
Anvith Thudi, Gabriel Deza, Varun Chandrasekaran, and Nicolas Papernot.
\newblock Unrolling sgd: Understanding factors influencing machine unlearning.
\newblock In \emph{2022 IEEE 7th European Symposium on Security and Privacy (EuroS\&P)}, pages 303--319. IEEE, 2022.

\bibitem[Tram{\`e}r et~al.(2016)Tram{\`e}r, Zhang, Juels, Reiter, and Ristenpart]{tramer2016stealing}
Florian Tram{\`e}r, Fan Zhang, Ari Juels, Michael~K Reiter, and Thomas Ristenpart.
\newblock Stealing machine learning models via prediction $\{$APIs$\}$.
\newblock In \emph{25th USENIX security symposium (USENIX Security 16)}, pages 601--618, 2016.

\bibitem[Veale et~al.(2018)Veale, Binns, and Edwards]{veale2018algorithms}
Michael Veale, Reuben Binns, and Lilian Edwards.
\newblock Algorithms that remember: model inversion attacks and data protection law.
\newblock \emph{Philosophical Transactions of the Royal Society A: Mathematical, Physical and Engineering Sciences}, 376\penalty0 (2133):\penalty0 20180083, 2018.

\bibitem[Wang et~al.(2023)Wang, Chen, Yuan, Zeng, Wong, and Yin]{wang2023kga}
Lingzhi Wang, Tong Chen, Wei Yuan, Xingshan Zeng, Kam-Fai Wong, and Hongzhi Yin.
\newblock Kga: A general machine unlearning framework based on knowledge gap alignment, 2023.

\bibitem[Warnecke et~al.(2021)Warnecke, Pirch, Wressnegger, and Rieck]{warnecke2021machine}
Alexander Warnecke, Lukas Pirch, Christian Wressnegger, and Konrad Rieck.
\newblock Machine unlearning of features and labels.
\newblock \emph{arXiv preprint arXiv:2108.11577}, 2021.

\bibitem[Yeom et~al.(2018)Yeom, Giacomelli, Fredrikson, and Jha]{yeom2018privacy}
Samuel Yeom, Irene Giacomelli, Matt Fredrikson, and Somesh Jha.
\newblock Privacy risk in machine learning: Analyzing the connection to overfitting.
\newblock In \emph{2018 IEEE 31st computer security foundations symposium (CSF)}, pages 268--282. IEEE, 2018.

\bibitem[Yevseiev et~al.(2021)Yevseiev, Laptiev, Lazarenko, Korchenko, and Manzhul]{yevseiev2021modeling}
Serhii Yevseiev, Oleksandr Laptiev, Sergii Lazarenko, Anna Korchenko, and Iryna Manzhul.
\newblock Modeling the protection of personal data from trust and the amount of information on social networks.
\newblock \emph{EUREKA: Physics and Engineering,(1)}, pages 24--31, 2021.

\bibitem[Yuan and Zhang(2022)]{yuan2022membership}
Xiaoyong Yuan and Lan Zhang.
\newblock Membership inference attacks and defenses in neural network pruning.
\newblock In \emph{31st USENIX Security Symposium (USENIX Security 22)}, pages 4561--4578, 2022.

\bibitem[Zbontar et~al.(2021)Zbontar, Jing, Misra, LeCun, and Deny]{pmlr-v139-zbontar21a}
Jure Zbontar, Li~Jing, Ishan Misra, Yann LeCun, and Stephane Deny.
\newblock Barlow twins: Self-supervised learning via redundancy reduction.
\newblock In Marina Meila and Tong Zhang, editors, \emph{Proceedings of the 38th International Conference on Machine Learning}, volume 139 of \emph{Proceedings of Machine Learning Research}, pages 12310--12320. PMLR, 18--24 Jul 2021.
\newblock URL \url{https://proceedings.mlr.press/v139/zbontar21a.html}.

\bibitem[Zhang et~al.(2023)Zhang, Lu, Zhang, Wang, and Li]{zhang2023machine}
Yongjing Zhang, Zhaobo Lu, Feng Zhang, Hao Wang, and Shaojing Li.
\newblock Machine unlearning by reversing the continual learning.
\newblock \emph{Applied Sciences}, 13\penalty0 (16):\penalty0 9341, 2023.

\end{thebibliography}
\bibliographystyle{plainnat}

\newpage
\appendix
\onecolumn


\end{document}